\documentclass[journal]{IEEEtran}
\usepackage{amsmath,amssymb}
\usepackage{booktabs}
\usepackage{graphicx}
\usepackage{microtype}
\usepackage{stfloats}
\usepackage{balance}
\usepackage[noadjust]{cite}
\usepackage{xcolor}
\usepackage{hyperref}
\begin{document}
\title{End-to-End Landsat Land Surface Temperature Reconstruction with AlphaEarth Embeddings}
\author{Shengjie~Liu,~\IEEEmembership{Member,~IEEE}%
\thanks{S. Liu was with the University of Southern California, Los Angeles, CA 90089, USA. Contact: skrisliu@gmail.com. } }
\markboth{Preprint Submitted to IEEE Geoscience and Remote Sensing Letters}{Liu: End-to-End Landsat LST Reconstruction}
\maketitle

\begin{abstract}
Land surface temperature (LST) data at Landsat's 30 m resolution is important for many applications, but cloud contamination prevents its spatially continuous observations. Here, we propose A$^2$E$^2$, an end-to-end U-Net that integrates annual temperature cycle (ATC), coarse LST from Earth system modeling, and AlphaEarth embeddings for Landsat LST reconstruction. The ATC supplies a physics-guided thermal prior, the coarse LST provides a temporal background, and the U-Net converts the embedding field into day-specific fine-scale correction using observed pixels. Experimental results on four seasonally distributed Landsat~8/9 scenes over New York City in 2022 show that A$^2$E$^2$ achieves $0.78\,\mathrm{K}$ mean absolute error and $1.01\,\mathrm{K}$ root mean squared error, reducing error by 60.1\% relative to baseline method without spatial correction and by 3.7\% relative to the same U-Net guided by annual mean spectral reflectance. Sensitivity analysis shows stable performance regardless of distance to clear pixels and percentage of blocked pixels. Independent validation with ECOSTRESS LST shows that Earth foundation model embeddings have better performance than annual mean spectral bands for LST reconstruction. 
\end{abstract}


{\textcolor{blue} {Final version: 
\url{https://doi.org/10.1109/lgrs.2026.3754195} }}

\section{Introduction}
High-resolution land surface temperature (LST) is essential for characterizing surface energy exchange and supports applications such as urban heat assessment, vegetation monitoring, drought analysis, and wildfire monitoring~\cite{liu2024racial, vlassova2016effects, wan2004using, rivera2026characterizing}. Landsat provides operational LST at $30\,\mathrm{m}$, enabling distinguishing neighborhood-scale thermal contrasts and land-cover transitions~\cite{malakar2018operational}. Thermal infrared retrievals, however, are obstructed by clouds, leading to large and irregular gaps in the LST products. Continuous LST at Landsat's 30 m resolution therefore depends on reconstructing LST over missing pixels~\cite{liu2025daily}. 

Spatial reconstruction methods estimate missing pixels from information available within the same acquisition~\cite{wu2021spatially}. Interpolation and kriging rely primarily on geographic proximity, whereas graph-based propagation can connect pixels with similar surface characteristics~\cite{rolland2024improving}. These methods are effective when gaps are small and clear observations adequately represent their surroundings. Their reliability decreases with extensive cloud patterns and for heterogeneous cities, where nearby pixels may correspond to thermally distinct materials and the center of a gap may be far from any observation. Spatial evidence alone cannot reliably reconstruct LST at all missing pixels. 

Spatiotemporal reconstruction addresses this limitation by introducing a temporal component. Recent methods commonly build upon the annual temperature cycle (ATC), which represents the seasonal trajectory through location-specific mean, amplitude, and phase parameters~\cite{zhu2022reconstruction, perez2023combined, yang2024annual}. ATC supplies a complete and physically interpretable prior, but it cannot represent the synoptic anomaly of a particular acquisition. This limitation is pronounced for Landsat because its nominal 16-day revisit and cloud contamination provide much sparser temporal support than daily coarse-resolution sensors like MODIS. Recent approaches therefore combine ATC with temporally complete reanalysis or coarse LST to constrain the acquisition-day state~\cite{liu2025daily}. Even after this temporal adjustment, fine-scale departures from the ATC-based estimate remain largely unresolved.

Traditional approaches model these residuals using manually selected variables and statistical learners~\cite{wu2021spatially}. Spectral bands, normalized difference vegetation index (NDVI), land cover, and topography have been used with linear regression, random forests, Gaussian processes, and ensemble models~\cite{zhao2020reconstruction, liu2025daily}. Such corrections can improve an ATC baseline, but their behavior depends on the chosen predictors, sampling scheme, and study region. Pointwise models also describe each location through an isolated feature vector rather than learning its surrounding spatial configuration directly. The central issue is thus not only which regressor to use, but how to represent similarity so that observed residuals can be transferred reliably into cloud-covered areas~\cite{jia2024advances}.

A useful feature space should place surfaces with comparable thermal behavior near one another even when they are geographically separated. Spectral reflectance provides direct information on surface composition, but individual bands or indices describe only partial land surface structure. Self-supervised Earth foundation models learn broader representations from large volumne of unlabeled archives~\cite{xiao2025foundation}. In particular, Google's AlphaEarth Foundation model provides a globally consistent 64-dimensional annual embedding field derived from multimodal data~\cite{brown2025alphaearth}, and has recently drawn attention in air pollutant and LST reconstruction~\cite{fang2026performance, leeburkhart2026geospatial}. The ready-to-use Earth foundation model embeddings offer a rich basis for measuring pixel similarity and may guide LST residual reconstruction with sparse observations.

Using satellite embeddings in LST reconstruction is not straightforward. Treating it as a simple regression problem overlooks local and long-range contextual information~\cite{peng2019end}. Methods with multiple steps may also complicate their integration within existing frameworks. An end-to-end framework based on U-Net offers a simple way to translate embeddings into a spatially coherent residual field~\cite{liu2025resolution}. This design remains physics-guided: ATC and coarse LST define the temporal background, so the network learns only the unresolved fine-scale correction from clear Landsat pixels rather than an unconstrained absolute temperature field. Although U-Nets have proved effective for remote-sensing gap filling, their end-to-end use driven by satellite embeddings for Landsat LST reconstruction remains uncommon. 

Motivated by this formulation, we propose A$^2$E$^2$ that integrates ATC prior, coarse LST from Earth system modeling (ESM), and AlphaEarth embeddings through a U-Net for LST reconstruction. The ATC and coarse LST establish a complete acquisition-specific background, while the U-Net transfers residual information from observed pixels to cloud-covered regions according to learned surface similarity. The main contributions of this work are twofold: (1) a method enabled by physics-guided U-Net and AlphaEarth embeddings that decomposes LST reconstruction into ATC, temporal background, and fine-scale spatial correction through convolutions; (2) the introduction of AlphaEarth embeddings to model spatial residuals that effectively captures multiscale context.   

\section{Method}
\subsection{Problem formulation}
Let $\Omega$ be the $H\times W$ Landsat grid, with $x=(i,j)\in\Omega$ indexing a $30\,\mathrm{m}$ pixel. The dataset contains $D$ acquisitions indexed by $t\in\mathcal{T}={1,\ldots,D}$. Let $d_t\in{1,\ldots,N}$ be the day of year (DOY) of acquisition $t$, where $N=365$ or 366. We write $Y_t(x)$ for Landsat LST in kelvin and collect all scenes as $Y\in\mathbb{R}^{D\times H\times W}$. The observation mask $M_t(x)$ equals one for valid LST observation and zero otherwise. 

Additional input data includes field $Y_t^c(x)$, the coarse complete LST from ESM, instantiated by ERA5 skin temperature at $0.25^\circ$ and bilinearly resampled to the Landsat grid. The static field $E(x)\in\mathbb{R}^{C}$ contains $C$ surface-description features at pixel $x$, either AlphaEarth embeddings or annual mean spectral reflectance.  

This setting differs from supervised learning, where paired samples define a direct mapping $f(x)\mapsto y$: the reconstructed LST here is not available as target. Moreover, the spatial resolution difference is large, being $0.25^\circ$ ($31\,\mathrm{km}$) for ERA5 to $30\,\mathrm{m}$ for Landsat, a thousandfold difference. It is also not a simple reconstruction task given the coarse LST and ATC as prior. We therefore treat the problem as prior-guided partial-field reconstruction. ATC and coarse LST supplie physics-guided seasonal prior and temporal background, and the feature field guides how residual information observed is extended across missing regions. 

\subsection{Baseline: ATC and coarse LST}
For location $x$ and acquisition $t$, the enhanced ATC (eATC) baseline is
\begin{equation}
 \begin{split}
 \widehat Y_t^{\mathrm{B}}(x)={}&T_0(x)
 +A(x)\cos\!\left[\omega\{d_t-\varphi(x)\}\right]\\
 &+\gamma(x)Y_t^c(x), \qquad \omega=\frac{2\pi}{N},
 \end{split}
 \label{eq:eatc}
\end{equation}
where $\widehat Y_t^{\mathrm{B}}(x)$ is the LST estimate; $T_0(x)$ is the annual mean; $A(x)$ and $\varphi(x)$ are seasonal amplitude and phase; and $\gamma(x)$ is a spatially varying coefficient on the coarse LST $Y_t^c(x)$.  

After \eqref{eq:eatc}, the remaining observed residual is
\begin{equation}
 R_t(x)=Y_t(x)-\widehat Y_t^{\mathrm{B}}(x),
 \qquad M_t^{\mathrm{tr}}(x)=1.
 \label{eq:residual-target}
\end{equation}
The residual $R_t(x)$ is the day-specific correction not explained by seasonal and temporal background terms, which is modeled through feature-guided learning. The final reconstruction is
\begin{equation}
 \widehat Y_t(x)=\widehat Y_t^{\mathrm{B}}(x)+\widehat R_t(x),
 \label{eq:decomp}
\end{equation}
where $\widehat R_t$ denotes the predicted residual field. If a day of year contains no valid observation, the complete eATC baseline remains the fallback estimate. This study focuses on days with valid observation only. 

\subsection{Proposed: U-Net guided by AlphaEarth}
In the proposed method, residuals are modeled through U-Net within one model that also jointly solves for eATC parameters, given feature field $E\in\mathbb{R}^{C\times H\times W}$. For AlphaEarth, $C=64$; for spectral bands, $C$ is the number of channels. The U-Net encoder aggregates local patterns over progressively broader neighborhoods, while the decoder combines this multiscale context with fine-scale information through skip connections. Let $Z_\theta(E)$ denote the resulting spatial feature map, where $\theta$ contains the shared encoder and decoder parameters. A date-specific output head then selects the residual associated with acquisition DOY $d_t$:
\begin{equation}
 \widehat R_t=W_{d_t}*Z_\theta(E)+b_{d_t},
 \label{eq:unet-output}
\end{equation}
where $W_{d_t}$ and $b_{d_t}$ are the convolutional weight and bias for DOY $d_t$, and $*$ denotes convolution.  

The proposed model, including U-Net ($\vartheta$) and eATC parameters ($T_0, A, \varphi, \gamma$), is jointly trained by minimizing masked $L_1$ loss over observed pixels:
\begin{equation}
 \begin{split}
 \mathcal{L}(\vartheta, T_0, A, \varphi, \gamma)={}&\frac{1}{|\mathcal{T}_+|}
 \sum_{t\in\mathcal{T}_+}\frac{1}{n_t^{\mathrm{tr}}}
 \sum_{x\in\Omega}\\[-0.2em]
 &M_t^{\mathrm{tr}}(x)
 \left|\widehat Y_t^{\mathrm{B}}(x)+\widehat R_t(x)-Y_t(x)\right|.
 \end{split}
 \label{eq:masked-l1}
\end{equation}

\subsection{Competitors}
Only the proposed method allows for joint optimization. Competitors such as Gaussian processes and classic vegetation-based correction require solving eATC parameters and extracting residuals $R_t$ first. Let $F(x)$ denote either the AlphaEarth embedding or annual mean spectral reflectance. The learned competitors use the same feature set
\begin{equation}
 q_t(x)=[F(x),\widetilde x_1,\widetilde x_2],
 \label{eq:gp-input}
\end{equation}
where $\widetilde x_1$ and $\widetilde x_2$ are normalized image coordinates.  

\paragraph{Classic vegetation-based spatial correction} 
The four-step ATC propagates observed corrections to missing pixels using manually defined similarity in geographic distance and NDVI~\cite{zhu2022reconstruction}. We apply this correction after the eATC baseline for a consistent residual comparison.

\paragraph{Sliding-window CNN}
The CNN applies a shared local sliding-window regressor to $q_t$ and predicts the residual pixel by pixel. Its fixed receptive field tests whether local context is sufficient without the U-Net's multiscale structure.

\paragraph{Gaussian-process models}
The DFGP~\cite{liu2024deep} maps $q_t(x)$ through $h_\eta$ to $z_t(x)=h_\eta(q_t(x))$ and learns residual covariance in this latent feature space:
\begin{equation}
 R_t(x)\sim\mathcal{GP}\!\left(
 \mu,\,
 k_\rho\{z_t(x),z_{t'}(x')\}
 \right).
 \label{eq:dfgp}
\end{equation}
Here, $\mu$ is the GP mean, $k_\rho$ is an RBF kernel with parameters $\rho$, and the paired arguments denote two acquisition--pixel samples. The DFGPs variant uses $q_t$ in the residual mean but restricts covariance to normalized geographic coordinates, thereby enforcing spatial smoothness instead of learned feature-space similarity. Both variants use their posterior means as $\widehat R_t$ and sparse variational inference for tractability~\cite{wilson2016deep}.

\section{Results and Analyses}
\subsection{Experimental setup}
The study domain covers New York City and its adjacent areas in Landsat paths/rows 013032 and 014032. The common grid contains $1635\times2116$ pixels at $30\,\mathrm{m}$. Of 60 daytime Landsat~8/9 Collection~2 LST scenes acquired in 2022, eight with less than 10\% valid coverage are removed, leaving $D=52$ acquisitions. Product quality information and finite-value tests define the clear-sky mask $M_t$. ERA5 skin temperature provides the coarse field $Y_t^c$. The spatial feature field $E$ is either the 64-channel AlphaEarth Satellite Embeddings in 2022 or annual mean Landsat spectral reflectance. These data are collected from Google Earth Engine. Experiments were carried out on a workstation with NVIDIA v100 16 GB and 128 GB RAM. 

\subsubsection{Mask-transfer validation}
We selected four dates with valid pixels ranging from $66.8\%$ to $98.8\%$ and with cloud masks transferred from different dates. This cloud pattern transfer technique allows us to evaluate the reconstruction performance in a realistic setting where the target scene is partially cloudy, while the donor scene provides a valid realistic cloud mask. The four target dates are 10 January, 15 April, 26 June, and 15 September, covering all four seasons, with corresponding donor dates of 13 August, 29 August, 13 July, and 28 July, respectively. Table~\ref{tab:holdouts} summarizes the split, whose four test masks contain 7,109,795 pixels as test pixels. 

Pooled MAE, RMSE, and $R^2$ are computed over their union, while per-scene MAE describes seasonal variation. Error reduction is reported relative to the pooled eATC MAE. The competitors include DFGP with modifications to the input features~\cite{liu2024deep}, a sliding window CNN modified for regression tasks~\cite{liu2024deep}, and the 4-step ATC~\cite{zhu2022reconstruction} enhanced with ERA5.

\begin{table}[t]
\caption{Mask-transfer split as a percentage of the full scene.}
\label{tab:holdouts}
\centering
\scriptsize
\begin{tabular*}{0.45\textwidth}{@{\extracolsep{\fill}} lcccccc }
\toprule
Date & DOY & Donor & Valid & Mask & Train & Test \\
\midrule
10 Jan & 10  & 13 Aug & 66.8\% & 39.5\% & 30.5\% & 36.3\% \\
15 Apr & 105 & 29 Aug & 98.3\% & 41.1\% & 40.4\% & 57.9\% \\
26 Jun & 177 & 13 Jul & 76.6\% & 37.8\% & 22.2\% & 54.4\% \\
15 Sep & 258 & 28 Jul & 98.8\% & 42.5\% & 41.9\% & 56.9\% \\
\bottomrule
\end{tabular*}
\end{table}

\subsubsection{Independent validation}
In a separate experiment, we obtained one ECOSTRESS LST on 9 Oct 2022 at 15:07 UTC for independent validation. The corresponding Landsat scene was obtained at 15:34 UTC. ECOSTRESS LST data were bilinearly resampled to Landsat resolution.

\subsection{Results under real-world cloud patterns}

\begin{table*}[t]
\caption{Results over Held-out test pixels on four dates across four seasons. Date columns are MAE (K); pooled metrics use all test pixels. Reduction is relative to the baseline eATC without spatial correction.}
\vspace{-0.75em}
\label{tab:results}
\centering
\small
\begin{tabular*}{0.9\textwidth}{@{\extracolsep{\fill}} llcccccccr }
\toprule
Method & Features & 10 Jan & 15 Apr & 26 Jun & 15 Sep & MAE & RMSE & $R^2$ & Reduction (\%) \\
\midrule
eATC & -- & 1.12 & 2.91 & 1.75 & 1.72 & 1.96 & 2.24 & 0.970 & -- \\
4-step ATC & -- & 0.91 & 1.01 & 1.51 & 0.70 & 1.04 & 1.34 & 0.989 & 46.9\% \\
DFGPs & spectral & 1.26 & 2.99 & 1.24 & 1.52 & 1.81 & 2.11 & 0.972 & 7.2\% \\
DFGPs & AlphaEarth & 1.28 & 2.85 & 1.30 & 1.46 & 1.78 & 2.07 & 0.974 & 9.1\% \\
CNN & spectral & 0.93 & 0.89 & 1.24 & 0.73 & 0.95 & 1.22 & 0.991 & 51.7\% \\
CNN & AlphaEarth & 1.04 & 1.66 & 1.56 & 1.21 & 1.40 & 1.68 & 0.984 & 28.4\% \\
DFGP & spectral & 0.70 & 1.07 & 1.15 & 0.76 & 0.94 & 1.18 & 0.992 & 52.1\% \\
DFGP & AlphaEarth & 0.83 & 1.04 & 1.60 & 0.96 & 1.13 & 1.41 & 0.988 & 42.3\% \\
A$^2$E$^2$ & spectral & 0.70 & \textbf{0.82} & \textbf{1.14} & 0.56 & 0.81 & 1.05 & 0.993 & 58.5\% \\
A$^2$E$^2$ & AlphaEarth & \textbf{0.60} & \textbf{0.82} & \textbf{1.14} & \textbf{0.50} & \textbf{0.78} & \textbf{1.01} & \textbf{0.994} & \textbf{60.1\%} \\
\bottomrule
\end{tabular*}
\vspace{-0.5em}
\end{table*}

Table~\ref{tab:results} reports summary results over the four dates across four seasons under real-world cloud patterns. The baseline without spatial correction (eATC) captures the annual range, with already fairly good performance given the introduction of ERA5 as temporal background and the per-pixel fitting design ($R^2=0.970$, RMSE=$2.24\,\mathrm{K}$). Yet, improvements are still needed, as spatial correction has not been applied, which lead to inferior performance over some days that deviated from the regional background, such as $2.91\,\mathrm{K}$ in April. The ATC prior and coarse temporal background cannot recover acquisition-specific fine structure alone.

Over all methods, the proposed A$^2$E$^2$ provides the lowest RMSE and MAE on all dates, regardless of season. With AlphaEarth, it attains $0.78\,\mathrm{K}$ MAE, $1.01\,\mathrm{K}$ RMSE, and $R^2=0.994$, a 60.1\% reduction from the baseline without spatial correction. January improves from 1.12 to $0.60\,\mathrm{K}$, April improves from $2.91$ to $0.82\,\mathrm{K}$, September reaches $0.50\,\mathrm{K}$, and June is most difficult, from 1.75 to $1.14\,\mathrm{K}$. 

AlphaEarth embeddings outperform spectral reflectance in the end-to-end U-Net, but not in the CNN or DFGP. This crossover identifies a representation--architecture interaction: the Earth foundation model embedding becomes useful when integrated across scales. The four-step ATC is stable but less accurate than A$^2$E$^2$, and the DFGPs variant is the least effective, indicating that geographic covariance alone is insufficient to transfer residuals across large cloud gaps.

\subsection{Sensitivity analysis}
\subsubsection{Distance to clear pixels}
\begin{figure}[htbp]
\centering
\includegraphics[width=0.9\linewidth]{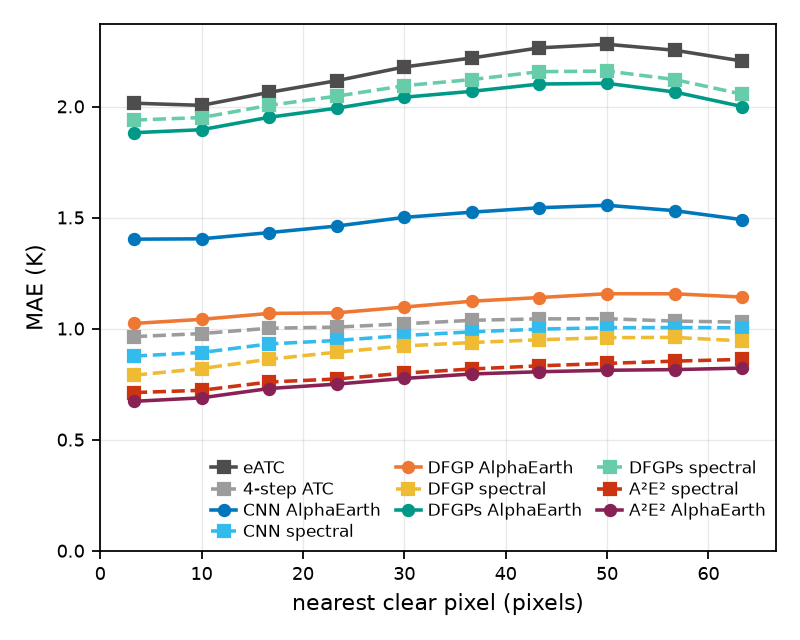}
\vspace{-1.5em}
\caption{Mean absolute error (MAE) vs. distance to the nearest clear pixel.}
\vspace{-0.5em}
\label{fig:distance}
\end{figure}

We examine the sensitivity of reconstruction error to distance from the nearest retained clear pixel and to the percentage of blocked pixels (Fig.~\ref{fig:distance}). The MAE is computed over all test pixels at a given distance or percentage. Although all methods degrade slightly with distance increasing, their ordering is stable. AlphaEarth A$^2$E$^2$ rises only from approximately $0.67$ to $0.82\,\mathrm{K}$ and remains below its spectral counterpart; the eATC baseline exceeds $2\,\mathrm{K}$ over most distances. The Earth foundation model embedding therefore remains useful inside large cloud gaps.

\subsubsection{Available observations}
\begin{figure}[htbp]
\centering
\includegraphics[width=0.9\linewidth]{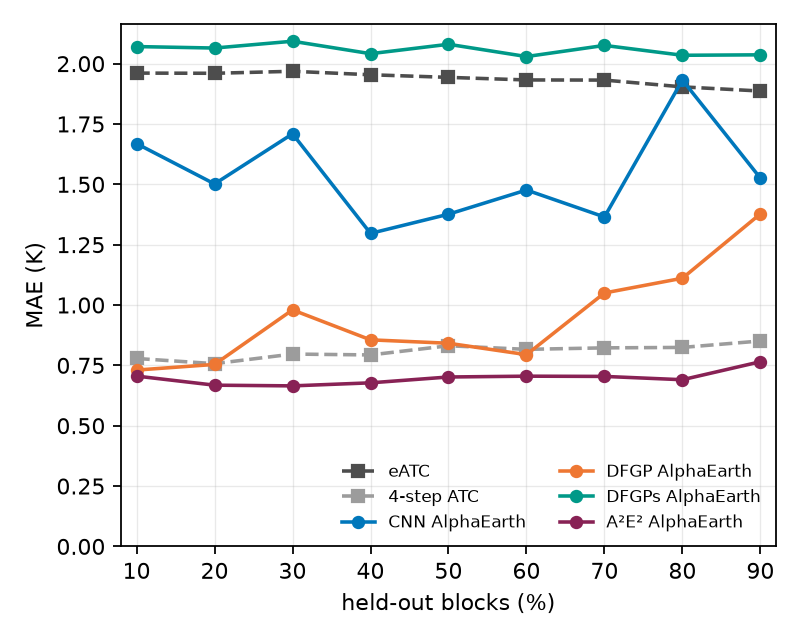}
\vspace{-1.5em}
\caption{Mean absolute error (MAE) vs. percentage of held-out $10\!\times\!10$ blocks.}
\vspace{-0.5em}
\label{fig:block}
\end{figure}

We designed a sensitivity analysis to test robustness to the percentage of available observations (Fig.~\ref{fig:block}). We divided the study domain into $10\!\times\!10$ blocks and randomly removed 10 of them, increasing from 10\% to 90\% removal. The remaining blocks are for training, and the removed blocks are for testing. The analysis ensures test pixels are spatially separated. Results show that A$^2$E$^2$ is robust to severe observation loss. Even with 90\% of the blocks removed, it maintains $0.76\,\mathrm{K}$ MAE. The four-step ATC is stable but less accurate; DFGP degrades beyond 60\% removal, and the local CNN reaches approximately $1.9\,\mathrm{K}$ at 80\%. Multiscale embedding guidance is robust to both distance and severe observation loss. The concern over reconstruction far away from observed pixels is largely resolved by the multiscale U-Net and the Earth foundation model embedding.

\subsubsection{Features}
Although AlphaEarth embeddings present superior performance, over some occasions it is not as good as spectral reflectance. Their comparison in Table~\ref{tab:results} shows that AlphaEarth embeddings outperform spectral reflectance with U-Net, but not with CNN or DFGP. AlphaEarth Embeddings have 64 channels, while spectral reflectance has 7 channels. Many of the machine learning methods, such as Gaussian processes, struggle with high-dimensional features, which may explain the performance difference. Although AlphaEarth's 64 channels provide rich information, redundancy exists. Future Earth foundation models may benefit from reducing the total dimensions needed to represent the Earth surface, or apply regularization so that each channel captures unique information.

\subsubsection{Runtime}
On a machine with NVIDIA v100 16 GB, over the NYC study area with 1635$\times$2116 pixels for all days in 2022, the proposed method finishes training and prediction in 126 s with AlphaEarth embeddings, and 118 s with spectral bands due to fewer input channels.

\subsection{Independent validation}
We conducted a separate experiment, holding out 9 October data (mask: 13 August, 59.9\% masked out), and validated it against an ECOSTRESS LST scene acquired 27 min earlier on the same date (Table~\ref{tab:result_ecostress}, Fig.~\ref{fig:maps}). Landsat and its reconstructions show a positive bias, consistent with its closer-to-midday acquisition time. The bias inherent to the data is unavoidable due to the observational timing difference. Among the reconstructions, methods paired with AlphaEarth embeddings overall have better performance than those paired with spectral bands, showing that AlphaEarth is better at capturing land surface characteristics.

\begin{table}[htbp]
\centering
\caption{Results validated against ECOSTRESS LST, 9 October 2022.}
\vspace{-0.5em}
\label{tab:result_ecostress}
\scalebox{1.0}[1.0]{
\begin{tabular}{l l c c c}
\toprule
Model / Data & Feature & RMSE (K)  & MAE (K) & Bias \\
\midrule
ATC+ERA5 & -- & 1.65 & 1.36 & 1.05 \\
4-step ATC & -- & 2.43 & 1.98 & 1.85 \\
DFGPs & AlphaEarth & 1.97 & 1.65 & 1.45 \\
DFGPs & Spectral & 1.97 & 1.68 & 1.51 \\
CNN & AlphaEarth & \underline{1.58} & \underline{1.23} & \underline{0.85} \\
CNN & Spectral & 1.75 & 1.46 & 1.20 \\
DFGP & AlphaEarth & 1.77 & 1.44 & 1.06 \\
DFGP & Spectral & 1.86 & 1.51 & 1.19 \\
A$^2$E$^2$ & AlphaEarth & \textbf{1.52} & \textbf{1.20} & \textbf{0.83} \\
A$^2$E$^2$ & Spectral & 1.66 & 1.34 & 0.98 \\
\bottomrule
\end{tabular} }
\end{table}
\vspace{-1.0em}

\subsection{Reconstructed LST maps}
\begin{figure*}[!t]
\centering
\includegraphics[width=.98\linewidth]{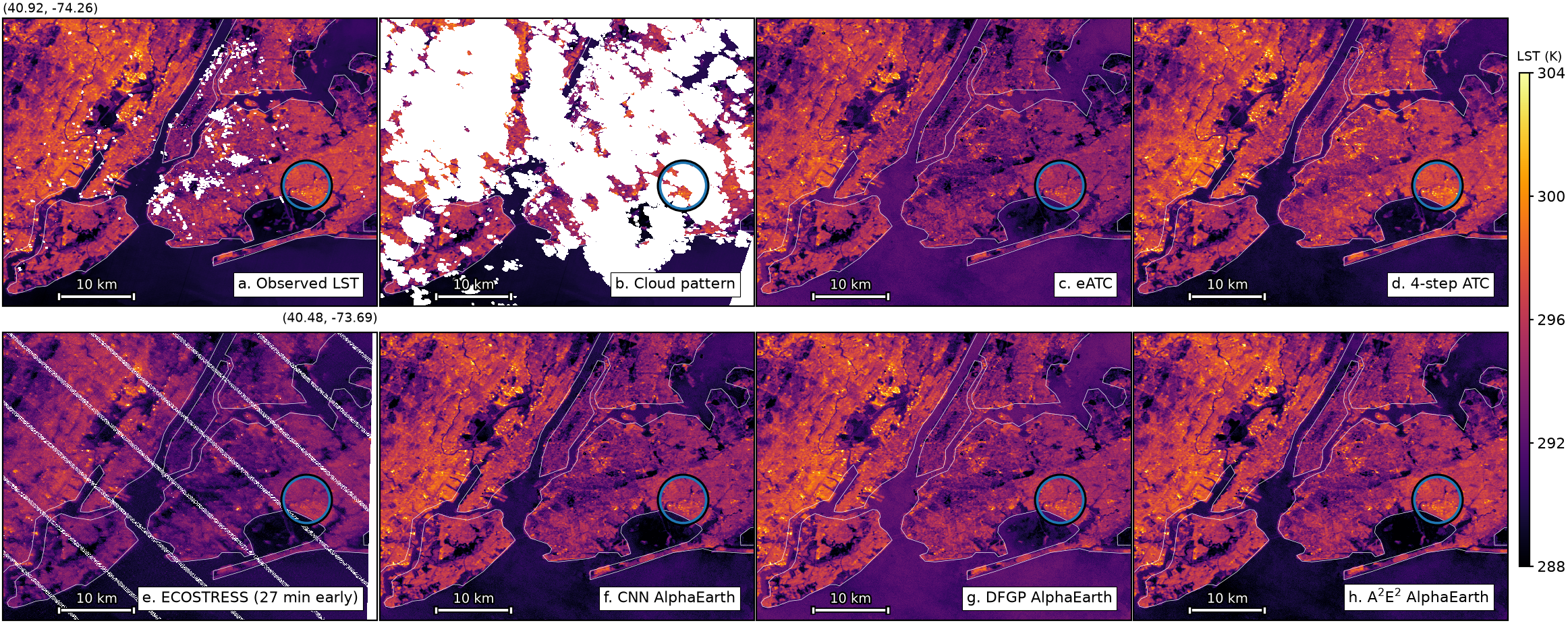}
\vspace{-0.5em}
\caption{LST reconstruction for 9 October 2022. Panels show the observed data, cloud pattern, the eATC baseline, four-step ATC, and ECOSTRESS observations, CNN, DFGP, and A$^2$E$^2$ reconstructions guided by AlphaEarth.}
\vspace{-0.5em}
\label{fig:maps}
\end{figure*}

We show the reconstructed LST maps in Fig.~\ref{fig:maps} for 9 October 2022. Only Landsat pixels were used in training, and the ECOSTRESS data acquired 27 min earlier were used for independent validation only. Overall, all methods show reasonable reconstructions, with A$^2$E$^2$ having slightly better agreement in capturing the LST contrast between deep water and urban areas (though the difference is visually marginal). Still, the reconstructed LST data show higher contrast compared with the 70 m ECOSTRESS data, demonstrating the necessity of Landsat's 30 m LST reconstruction.

\section{Conclusion}
In this study, we propose an AlphaEarth embeddings-based method for Landsat land surface temperature (LST) reconstrution. The method is based on a U-Net encoded with annual temperautre cycle (ATC) prior, coarse LST from Earth system models (ESM) as temporal background, and convolutions over AlphaEarth embeddings to guide the day-specific fine-scale spatial correction. Over a series of experiments, results show that the proposed A$^2$E$^2$ reduced mean absolute error (MAE) from $1.96$ to $0.78\,\mathrm{K}$ relative to the baseline without spatial correction over scenes across four seasons. Sensitivity analysis shows robustness with large data gaps. Independent validation with clear-sky ECOSTRESS LST shows that AlphaEarth embeddings instead of spectral bands are more suitable for LST reconstruction across all methods. For future work, predicting ATC parameters from embedding fields may further utlize their rich representation power, and given the availability of ESM data and AlphaEarth embeddings, the proposed method can be generliazed to other global cities.   

\balance
\bibliographystyle{IEEEtran}
\bibliography{refs}

\end{document}